\documentclass{IOS-Book-Article}

\usepackage{mathptmx}
\usepackage{import}
\usepackage[utf8]{inputenc}
\usepackage{tabularx}
\usepackage{graphicx}
\usepackage{xcolor}
\usepackage{pgfplots}
\pgfplotsset{compat=1.14}
\usepackage{tikz}
\usepackage{comment}

\def\hb{\hbox to 10.7 cm{}}

\begin{document}

\pagestyle{headings}
\def\thepage{}

\begin{frontmatter}              

\title{Named-Entity Linking Using Deep Learning For Legal Documents: A Transfer Learning Approach}

\markboth{}{September 2018\hb}

\author{\fnms{Ahmed} \snm{Elnaggar}%
\thanks{Corresponding Author: Ahmed Elnaggar, Software Engineering for Business Information Systems, Boltzmannstr. 3, 85748 Garching bei München, Germany; E-mail: ahmed.elnaggar@tum.de.}},
\author[A]{\fnms{Robin} \snm{Otto}},
and
\author[A]{\fnms{Florian} \snm{Matthes}}

\runningauthor{Ahmed Elnaggaer}
\address[A]{Software Engineering for Business Information Systems, Technische Universität München, Germany}

\begin{abstract}
In the legal domain it is important to differentiate between words in general, and afterwards to link the occurrences of the same entities. The topic to solve these challenges is called Named-Entity Linking (NEL). Current supervised neural networks designed for NEL use publicly available datasets for training and testing. However, this paper focuses especially on the aspect of applying transfer learning approach using networks trained for NEL to legal documents. Experiments show consistent improvement in the legal datasets that were created from the European Union law in the scope of this research. Using transfer learning approach, we reached F1-score of  98.90\% and 98.01\% on the legal small and large test dataset.

\end{abstract}

\end{frontmatter}
\markboth{September 2018\hb}{September 2018\hb}

\section{Introduction}
About 30 years ago, organizations started to automate different processes in their enterprise with the help of computers.
It started with applications whose absence, from today's perspective, is inconceivable for almost every single organization in the world.
An example for such a necessity was developing websites to create an interface between the customers and the enterprises.
This usage of digitized data has evolved into countless areas, dependent on the individual needs of every company~\cite{heilig2017analysis}.

Today's goals are naturally more advanced.
A large topic in state of the art research is to make computers able to understand natural language.
This process, called natural language processing, is extremely difficult due to the complexity of human language.
Numerous different approaches that achieve extremely well on different topics can be found online.
Though, this does not apply to the legal domain, as its content is more sensitive and less data to experiment with is publicly available.
Approaching the legal domain and its formulations in the context of natural language processing is still in a very early stage.
Therefore, any result that can be extracted from research in this segment is highly interesting~\cite{erdelez1997legal}.

One of these challenges in the legal domain is NEL, where we need to differentiate between words in general and afterwards to link the occurrences of the same entities. 
For example, in ``Paris killed the victim, while she was angry``, the challenge would be to associate the word ``Paris`` with a person not a city, as well as to link the occurrences of ``Paris`` and ``she`` to the same entity.
Current supervised neural networks designed for NEL use publicly available datasets for training and testing. They address the problem of automated information extraction in general and try to maximize the accuracy and speed of these networks. However, in this paper, we focus on applying NEL using transfer learning approach specifically in the legal documents.

Through our work, we will try to answer the following three research questions:
\begin{itemize}
\item \textit{Is the use of transfer learning with NEL beneficial in the legal domain?}
\item \textit{Which technique of transfer learning suits best?}
\item \textit{What kind of existing approach should be used for transfer learning in NEL problem?}
\end{itemize}

\section{Related Work}
\subsection{Transfer Learning}
Sawada et al. \cite{sawada2017all} point out that in the state of the art in supervised transfer learning \cite{pan2010survey,hu2015deep, rosenstein2005transfer}, most researchers take a similar approach.
First, a base model is constructed that has been trained on source domain data using a source cost function. Then, a second model is implemented based on the target domain data.
This time, the hidden layers from the original model are used as initial values and the output layer is replaced.
Additionally a second cost function for the new model is applied.
They furthermore state that this approach in fact outperforms non-transfer learning when the source and target domain are similar.
The convolutional neural network is trained on a source task with an original classifier.
Everything is then applied to a new input, the developers only changed the classifier.
Sawada et al. \cite{sawada2017all} try to prevent the overfitting by also taking the output layer into account when applying transfer learning.
For the transferred output layer, a new cost function is calculated.
The difference is that this new cost function is not constructed based on the target domain, it rather consists of two cost functions that get linked by evaluating the relationship between source and target data.
In the case of \cite{sawada2017all} the source data originated from the MNIST and CIFAR-10 images datasets.
The target domain dataset was generated from two-dimensional electrophoresis images, that can indicate diseases of patients.
The resulting model of applying deep neural networks to the images achieved an accuracy of over 90\%.

\subsection{NEL with Transfer Learning}
Over the last years, multiple studies focused on the task of NEL as a subtask to natural language processing.
Avirup et al.~\cite{sil2012linking} point out that most of the approaches use Wikipedia as a knowledge base in their work.
Especially when creating an NEL system for a certain domain, expert knowledge from that area is a huge advantage.
Only a small part of that expert knowledge, which would be needed to reach higher accuracies in many cases, can be found in Wikipedia.
Thus, Avirup et al.~\cite{sil2012linking} present another approach called Open-DB NED.
They use a domain independent, database-driven feature generation meaning the features of the algorithm will be adapted depending on the current knowledge base.
This approach defines different count functions as features which can be used independent of the domain and the knowledge base.
Their counts are based on the attributes of the different entities as well as the similarities between entities.~\cite{sil2012linking} 

Another approach provides a language independent entity linking.
Radu et al.~\cite{sil2017one} derive features, that are able to discriminate between entities across languages.
They state that all their features base on a similarity measurement between the context of the mention and the correct entity.
The presented results in~\cite{sil2017one} show, that their system trained in English, still outperforms other state of the art systems when tested in Chinese and Spanish.

\subsection{NEL in Legal Domain}
Making a machine learning algorithm learn domain specific knowledge, it needs a huge amount of training data.
In the legal domain, only few corpora exist, most of them relatively small in size.
The scarcity of data is one of many reasons why NEL in terms of law has not gotten much attention compared to other domains.
The only notable approach that is presented here is a system for named-entity recognition and -linking for legal texts.
The authors Cardellino et al.~\cite{cardellino2017low} implemented a system for named-entity recognition, classification and linking.
While they focus on a legal corpus for the recognition and classification task, entity linking was only evaluated on a Wikipedia corpus.
Therefore, they also did not entirely tackle the problem of NEL in the legal domain.

\section{Datasets}
Table \ref{tab:ed_datasets} shows statistics of the most popular name entity linking public datasets, as well as the dataset we have created for the legal domain. A summary description of these datasets is explained below.
\begin{table}[!ht]
\centering
\begin{tabular}{|l|l|l|l|l|}
\hline
Dataset         & Number Mentions & Number Documents & Gold Recall \\ \hline
AIDA-train      & 18,848           & 946              & 100\%       \\ \hline
AIDA-A (valid)  & 4,791            & 216              & 96.9\%      \\ \hline
AIDA-B (test)   & 4,485            & 231              & 98.2\%      \\ \hline
MSNBC           & 656             & 20               & 98.5\%      \\ \hline
AQUAINT         & 727             & 50               & 94.2\%      \\ \hline
ACE2004         & 257             & 36               & 90.6\%      \\ \hline
WNED-CWEB       & 11,154           & 320              & 91.1\%      \\ \hline
WNED-WIKI       & 6,821            & 320              & 92\%        \\ \hline
EURLEX-train 1k & 1,853            & 1,118             & 87\%        \\ \hline
EURLEX-test 1k  & 333             & 185              & 87\%        \\ \hline
EURLEX-train 20k& 33,937           & 17,352            & 87\%        \\ \hline
EURLEX-test 20k & 11,674           & 4,580             & 87\%        \\ \hline
\end{tabular}
\caption{Entity Disambiguation Datasets. \textit{Gold Recall} represents the percentage of mentions which have the ground truth entity in their respective candidate set. This table was adapted from~\cite{ganea2017deep}.}
\label{tab:ed_datasets}
\end{table}
\subsection{Public Benchmark Datasets}
The following datasets are used in state of the art NEL systems to measure the performance of the developed algorithms.
They are publicly available and mostly manually annotated.

\begin{enumerate}
\item AIDA-CoNLL. \\
This dataset is widely used for public benchmarks on the entity disambiguation task.
In terms of entity disambiguation, the AIDA corpus is the biggest manually annotated dataset available.
\item WNED~\footnote{http://lemurproject.org/clueweb09/FACC1/}. \\
This is a collection of datasets automatically created from the Wikipedia corpus.
Since they are generated automatically and hence less trustworthy, they are mostly used for testing a NEL system, not for training it.
\item MSNBC \& ACE2004. \\
All three of these datasets are taken from different news corpora, all written in English.
The news datasets are small in comparison to the AIDA-CoNLL datasets but are also manually annotated.
\end{enumerate}

\subsection{EUR-Lex Dataset}
\label{sub:create_eurlex}
This dataset was created from the corpus publicly accessible on a sub-page of the official homepage of the European Union\footnote{http://eur-lex.europa.eu/homepage.html}.
The following steps were involved in the extraction of the necessary data:
\begin{enumerate}
\item Download documents in HTML format from the EUR-Lex website and extract the content into separate text files.
\item Load content from text files and send them to the API of \textit{Dandelion's} NEL system.~\footnote{https://dandelion.eu/}
\item Store response in a format similar to the format of the WNED datasets.
\end{enumerate}
After the generation of the dataset, we divided it into two parts.
One contains the annotations of all documents that were processed from the EUR-Lex corpus (EURLEX 20k), the other contains annotations from only a restricted set of approximately 1,000 documents (EURLEX 1k).

In this research, we only focus on the Aida dataset as a source for transfer learning. Simply, because it is the largest manually annotated datset exist for named entity linking. Table \ref{tab:dataset_sim} shows the percentage of entities similarities between Aida dataset and our generated legel dataset.
\begin{table}[!h]
\centering
\begin{tabular}{|c|c|c|c|c|}
\hline
\% of similar entities & Aida-train & Aida-test & EUR-Lex-train \\ \hline
Aida-test              & 53         & 100       & -             \\ \hline
EUR-Lex-train          & 7          & 4         & 100           \\ \hline
EUR-Lex-test           & 20         & 13        & 70            \\ \hline
\end{tabular}
\caption{Similarities between the datasets.}
\label{tab:dataset_sim}
\end{table}
\section{Methodology and approach}
\subsection{Choose a NEL System}
Transfer learning focuses on re-using a model that has been created for and trained on another dataset, whose domain is in some sort similar to the new dataset.
Hence, the focus of the work is to find a state of the art deep NEL system.
Afterwards, the goal is to apply transfer learning to that system to be able to re-use it for another dataset.

This section presents how such a system has been chosen.
First and foremost, the machine learning metrics \textit{accuracy} and \textit{F1 score} played an important role in choosing an adequate deep NEL system.
Searching in the pool of state of the art NEL models, with the criteria defined above, leads to numerous projects that have been made open-source.
Refining the search with the machine learning metrics and the restriction to deep learning architectures, the number of possible systems reduced drastically.
Only one competitor satisfied all requirements to be adequate for the scope of this work.

\subsection{Deep Learning Architecture}
\label{sub:dl_archictecutre}
The paper used for this approach~\cite{ganea2017deep} presents a new deep learning model for named-entity disambiguation.
According to the authors, it combines aspects from two research perspectives.
The approach combines benefits of deep learning with traditional machine learning experiments.
Their developed architecture achieves F1 scores of state-of-the-art systems and even outperforms many approaches.
\subsubsection{Overview}
The input for the architecture implemented in~\cite{ganea2017deep} needs to be provided in the following format.
\begin{itemize}
\item Document name: The name of the document a mention occurred in.
\item Mention: The mention that has to be linked to an entity.
\item Context: The left and right context window of 100 words around the mention.
\item Candidates: The set of entity candidates for the respective mention.
\item Ground truth: the correct entity from the candidate set for the mention.
\end{itemize}
The model consists of the following three different parts.
\begin{itemize}
\item Entity Embedding.
The entity embedding contains the semantic meaning of entities.
For each entity, the embedding is inherited from pre-trained word embeddings.
In this approach, the authors embedded both, words and entities, together.
This gives the opportunity to calculate geometric similarities between words that surround a certain entity and the entity itself.
\item Local Model with Neural Attention.
The authors assume that only some context words are important for the disambiguation step of a mention.
For each entity with the respective context words, a context score is calculated, ranking the importance of each context word with respect to the current entity. 
To remove some noise, the algorithm only considers the top 20 words from the local context.
The goal for the local model is then to find a function that assigns a higher score to the correct entity that to the other entity candidates. 
\item Collective Disambiguation.
To put it all together, the authors propose to use a deep learning architecture.
This architecture resolves ambiguous mentions using a conditional random field with parametrized potentials.
\end{itemize}

\subsubsection{Experiments}
\label{sub:djed_experiments}
The authors from~\cite{ganea2017deep} trained, validated and tested their model on the public benchmark dataset AIDA-CoNLL.
AIDA-A,the validation set, was used for the adaption of the learning rate of the algorithm.
Amongst other datasets, automatically generated from different news corpora, they tested their model on AIDA-B, which is the denoted test dataset from the AIDA-CoNLL collection.
As described in~\cite{ganea2017deep} the performance peak with respect to the AIDA-B dataset was 92.2\%.
This result competes with most and even outperforms some of the systems that account for state of the art results.
We use their best result as a baseline for this paper.

\subsection{Apply Transfer Learning}
\begin{figure}[!htph]
\includegraphics[width=0.4\textwidth]{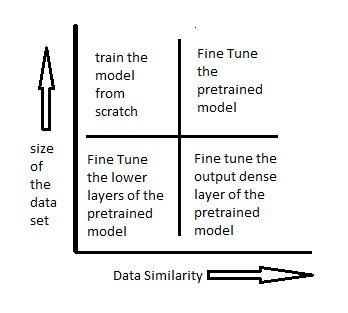}
\centering
\caption{The typical transfer learning scenarios}
\label{fig:TL_scenarios}
\end{figure}

Figure~\ref{fig:TL_scenarios}\footnote{https://data-flair.training/blogs/transfer-learning/} depicts the four possible scenarios for transfer learning.
This paper focuses on the high data similarity scenarios.
The necessary steps for \textit{fine tune the pretrained model} and \textit{fine tune the output layer of the pretrained model} are explained in the following subsection.

\subsubsection{Training}
First, we needed to include all new datasets in the entity embedding training.
Without this step, the network will later maybe predict the correct entity from the candidate set, but cannot link it to a knowledge base entry as it has not been included in the embedding.
The entity embedding serves as a lookup map for entities.
\begin{figure}[!ht]
\includegraphics[width=\textwidth]{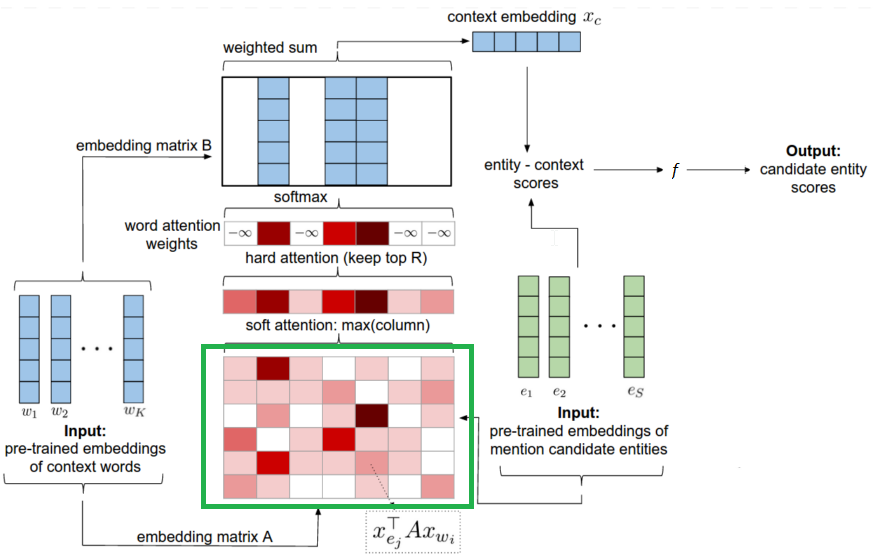}
\centering
\caption{The neural network used for local context scores of entities adapted from~\cite{ganea2017deep}.}
\label{fig:local_model}
\end{figure}
Figure~\ref{fig:local_model} depicts the architecture of the local model implemented by Ganea et al.~\cite{ganea2017deep} and indicates (green) what part had to be replaced by the new entity embedding.
As stated by the authors, function \textit{f} is responsible for determining the context scores.
They furthermore indicate that a flexible choice of \textit{f} is important for the performance of the model.
Hence, they used a neural network for \textit{f} consisting of two fully connected layers of 100 hidden units and ReLU non-linearities. 

After the creation of that entity embedding, the focus was to create baseline models.
With these baseline models, we were able to compare the transfer learning results to results achieved by training on a single dataset.
Afterwards, the same procedure was applied to the EUR-Lex dataset.\\
First, the legal dataset was divided into a train and test set, similar to AIDA-CoNLL.
Table~\ref{tab:dataset_sim} shows the overlapping percentage of entities in the respective train and test sets.
The overlap of Aida-train and test is 53\% while EUR-Lex-train contains 70\% of the entities included in the test set.
The gap of 17\% comparing the two corpora is not optimal but the best that could be achieved without involving a huge amount of manual work.

\subsubsection{Testing}
We employed F1 score and standard accuracy in the testing phase to draw conclusions on the performance of an approach.
Generally, we tested in two different topics: single training and transfer learning.
Transfer learning was applied on models trained on the EUR-Lex and AIDA-CoNLL.
This makes it possible to conclude whether fine tuning a model trained on a generalized dataset outperforms training on a niche dataset and fine tuning with the help of a well generalized one.
\section{Experimental Study and Performance Evaluation}
\subsection{Experimental Setup (Hardware \& Metrics)}
We used a machine contains a NVIDIA GeForce GTX TITAN X graphics card with 12 GB memory.
Furthermore, it has an Intel Core i7-5820k
The RAM size of the machine learning computer is 16 GB. With this setup, the average training time for the involved datasets was 60 hours.
The training process was automatically stopped using the early stop approach to solve over-fitting problem.

The following matrices were used to measure the performance of the results:
\begin{itemize}
\item \textit{Precision}.
Precision indicates how many of the classified mentions were correctly classified.
\item \textit{Recall}.
Recall represents the number of examples with ground truth assignment that have been guessed by the model.
\item \textit{F1 score}.
The F1 score is the harmonic average of \textit{precision} and \textit{recall}.
Therefore, it acts as a good performance measurement in classification problems.
Thus, it is the most common metric in NEL and the generally accepted measurement for comparing the performances of different systems.
\end{itemize}

\subsection{Results}
Table \ref{tab:tl_summary} summarize the results of both single training and transfer learning. In the following section we describe them in details.

\begin{table}[!ht]
\centering
\begin{tabular}{c|c|c|}
\cline{2-3}
                                        & Single Training 						 & Transfer Learning              \\ \hline
\multicolumn{1}{|l|}{AIDA-train}        & 92.36\%         						 & {\color[HTML]{009901} \textbf{93.41\%}} \\ \hline
\multicolumn{1}{|l|}{AIDA-test}            & {\color[HTML]{009901} \textbf{90.1\%} }         & 88.8\%  \\ \hline
\multicolumn{1}{|l|}{EUR-Lex-train 1k}  & 99.02\%         						 & {\color[HTML]{009901} \textbf{99.73\%}} \\ \hline
\multicolumn{1}{|l|}{EUR-Lex-train 20k} & 98.34\%         						 & {\color[HTML]{009901} \textbf{98.49\%}} \\ \hline
\multicolumn{1}{|l|}{EUR-Lex-test 1k}   & 98.29\%        						 & {\color[HTML]{009901} \textbf{98.90\%}} \\ \hline
\multicolumn{1}{|l|}{EUR-Lex-test 20k}  & {\color[HTML]{009901} \textbf{98.01}}\%        						 & {\color[HTML]{009901} \textbf{98.01}}\%                        \\ \hline
\end{tabular}
\caption{F1 score comparison between all experiments.}
\label{tab:tl_summary}
\end{table}

\subsubsection{Single Training}
First, we want to analyze the results of the single training on the legal dataset.
The achieved F1 score exceeds 97.5\% after the first few epochs.
This indicates a fast convergence towards the dataset.
The network's weights seem to adapt rapidly in direction of the EUR-Lex corpus.
We assume there are multiple reasons for that behavior.
First, our EUR-Lex dataset contains almost 34,000 entries of entity-mention pairs.
These entries consist of 4,251 different pairs, where 83 of these pairs occur with a high frequency.
These numbers indicate that there is little variety of entities in the legal corpus.
Hence, we assume the network can easily learn to predict the correct entity for a mention due to repetition.
The final F1 score for EUR-Lex 1k and  EUR-Lex 20k test datasets are 98.29\% and 98.01\%.

\subsubsection{Transfer Learning}
Three different approaches are taken into account for judging on the success of transfer learning in a deep NEL system.
\begin{itemize}
\item AIDA-CoNLL $\rightarrow$ EUR-Lex 1k. \\
Here, the network immediately adapts its weights towards the EUR-Lex corpus in the first epochs.
The performance starts with an F1 score of 99\% and improves up to 99.73\% on training datset, while the test datasets reasched F1 score of 98.90\%.
Compared to single training on the EUR-Lex dataset, the achieved accuracies and F1 score were close to 99\%.
\item AIDA-CoNLL $\rightarrow$ EUR-Lex 20k. \\
This approach indicates the same conclusions as the experiment above.
It can be observed that the F1 score increased for the EUR-Lex 20k train dataset when employing transfer learning. However, the F1 score for the test dataset remained the same.
\item EUR-Lex $\rightarrow$ AIDA-CoNLL. \\
This approach produced promising and interesting results.
Although it started with a bias towards the EUR-Lex 20k dataset, which is bigger compared to the AIDA corpus, the network was able to learn the better generalized dataset with decent performance.
With a F1 score of 93.41\% after transfer learning, compared to 92.36\% in single training, it outperforms the original approach of Ganea et al.~\cite{ganea2017deep} with respect to the AIDA-train dataset.
Despite the fact that the F1 score increased when fine-tuning the model, it has to be noted that the networks performance decreases on the AIDA-test dataset by 1.3\% from 90.1\% to 88.8\%.
This indicates that the network could be overfitting on the AIDA-train dataset in this case.
Interpreting the above observations, it is required sometimes to changing the transfer learning scenarios as mentioned in Figure \ref{fig:TL_scenarios} or apply regularization techniques when using transfer learning to prevent over-fitting.   
\end{itemize}

\subsection{Discussion}
We want to conclude and discuss our results by answering the research questions motivating this work:
\begin{enumerate}

\item \textit{Is the use of transfer learning with NEL beneficial in the legal domain?} \\
By having a closer look on the final results, transfer learning for the EUR-Lex resulted in an improved F1 score compared to the single training experiment.
Unexpectedly, we even observed an improvement on the AIDA-CoNLL training dataset when employing transfer learning.
Although in this case the test F1 score dropped compared to the single training, we are confident that by applying and testing other transfer learning scenarios and applying more regularization techniques there is even more room for improvement. 
Finally, we can conclude that transfer learning was beneficial in the legal domain.
Furthermore, the observations even imply that the performance of NEL systems can improve generally through the use transfer learning.

\item \textit{Which technique of transfer learning suits best?} \\
The two involved datasets, AIDA-CoNLL and EUR-Lex, are both rich in data and have high similarity with respect to the high frequent entities.
Prior research suggests to employ the technique of fine tuning a network with the new dataset without exchanging any layer.
This is justified by indicating that if the datasets are similar, the network only needs to adapt its weights slightly towards the new dataset.
For this reason, we decided that the fine tuning technique is the most promising in this work.

\item \textit{What kind of existing approach should be used for transfer learning in named entity linking problem?} \\
Since NEL itself is a complicated task, not many approaches use deep learning to solve it.
To our best knowledge we couldn't find any related work coping with the exact same problem.
The model from~\cite{ganea2017deep} fulfilled the objective performance criteria and outperform most approaches in the state of the art.

\end{enumerate}

\section{Conclusion}

Transfer learning is a promising approach that could the way we solve problems in the legal domain, by transferring knowledge from large datasets to small datasets. We investigated the transfer learning capabilities on NEL for the European Union legal dataset. 
We proved that transfer learning on NEL can be useful for in the legal domain. Of course, the model architecture, transfer learning scenario are important factors to determine the performance. We think, it is an effective method to solve dastat scarcity problem in the legal domain. Using this approach, we might enhance the performance of most of all current state of the art results in the legal domain. This research could be a base for further research on the usage of transfer learning approach in the legal domain. However, more experiments are needed to test it in other problems.

\section{Acknowledgements}
We gratefully acknowledge the hardware support from iteratec GmbH\footnote{https://www.iteratec.de/} for this research.

\begin{bibliography}{bibliography/references.bib}
\bibliographystyle{plain}
\end{bibliography}
\end{document}